\documentclass[sigconf,authorversion,nonacm]{acmart} 

\usepackage{bm}

\newcommand{\etal}{\textit{et al}. }
\newcommand{\ie}{\textit{i}.\textit{e}.}
\newcommand{\eg}{\textit{e}.\textit{g}.}

\usepackage{mathtools}

\DeclarePairedDelimiter\floor{\lfloor}{\rfloor}

\hypersetup{
  colorlinks, linkcolor=red
}

\definecolor{myyellow}{RGB}{255, 192, 0}
\definecolor{mygreen}{RGB}{0, 176, 80}

\copyrightyear{2021}
\acmYear{2021}
\setcopyright{acmlicensed}
\acmConference[MM '21]{Proceedings of the 29th ACM International Conference on Multimedia}{October 20--24, 2021}{Virtual Event, China}
\acmBooktitle{Proceedings of the 29th ACM International Conference on Multimedia (MM '21), October 20--24, 2021, Virtual Event, China}
\acmPrice{15.00}
\acmDOI{10.1145/3474085.3475310}
\acmISBN{978-1-4503-8651-7/21/10}

\acmSubmissionID{819}

\begin{document}
\fancyhead{}

\title{Multiview Detection with Shadow Transformer\\ (and View-Coherent Data Augmentation)}


\author{Yunzhong Hou}
\affiliation{%
  \institution{Australian National University}
  \streetaddress{ACT 2600}
  \city{Canberra}
  \state{ACT 2600}
  \country{Australia}}
\email{yunzhong.hou@anu.edu.au}

\author{Liang Zheng}
\authornote{Corresponding author}
\affiliation{%
  \institution{Australian National University}
  \streetaddress{ACT 2600}
  \city{Canberra}
  \state{ACT 2600}
  \country{Australia}}
\email{liang.zheng@anu.edu.au}

\renewcommand{\shortauthors}{Trovato and Tobin, et al.}

\begin{abstract}

Multiview detection incorporates multiple camera views to deal with occlusions, and its central problem is multiview aggregation. Given feature map projections from multiple views onto a common ground plane, the state-of-the-art method addresses this problem via convolution, which applies the same calculation regardless of object locations. However, such translation-invariant behaviors might not be the best choice, as object features undergo various projection distortions according to their positions and cameras. In this paper, we propose a novel multiview detector, MVDeTr, that adopts a newly introduced shadow transformer to aggregate multiview information. Unlike convolutions, shadow transformer attends differently at different positions and cameras to deal with various shadow-like distortions. We propose an effective training scheme that includes a new view-coherent data augmentation method, which applies random augmentations while maintaining multiview consistency. On two multiview detection benchmarks, we report new state-of-the-art accuracy with the proposed system. Code is available at \url{https://github.com/hou-yz/MVDeTr}.



\end{abstract}




\maketitle

\section{Introduction}

\begin{figure}[h]
  \centering
  \includegraphics[width=\linewidth]{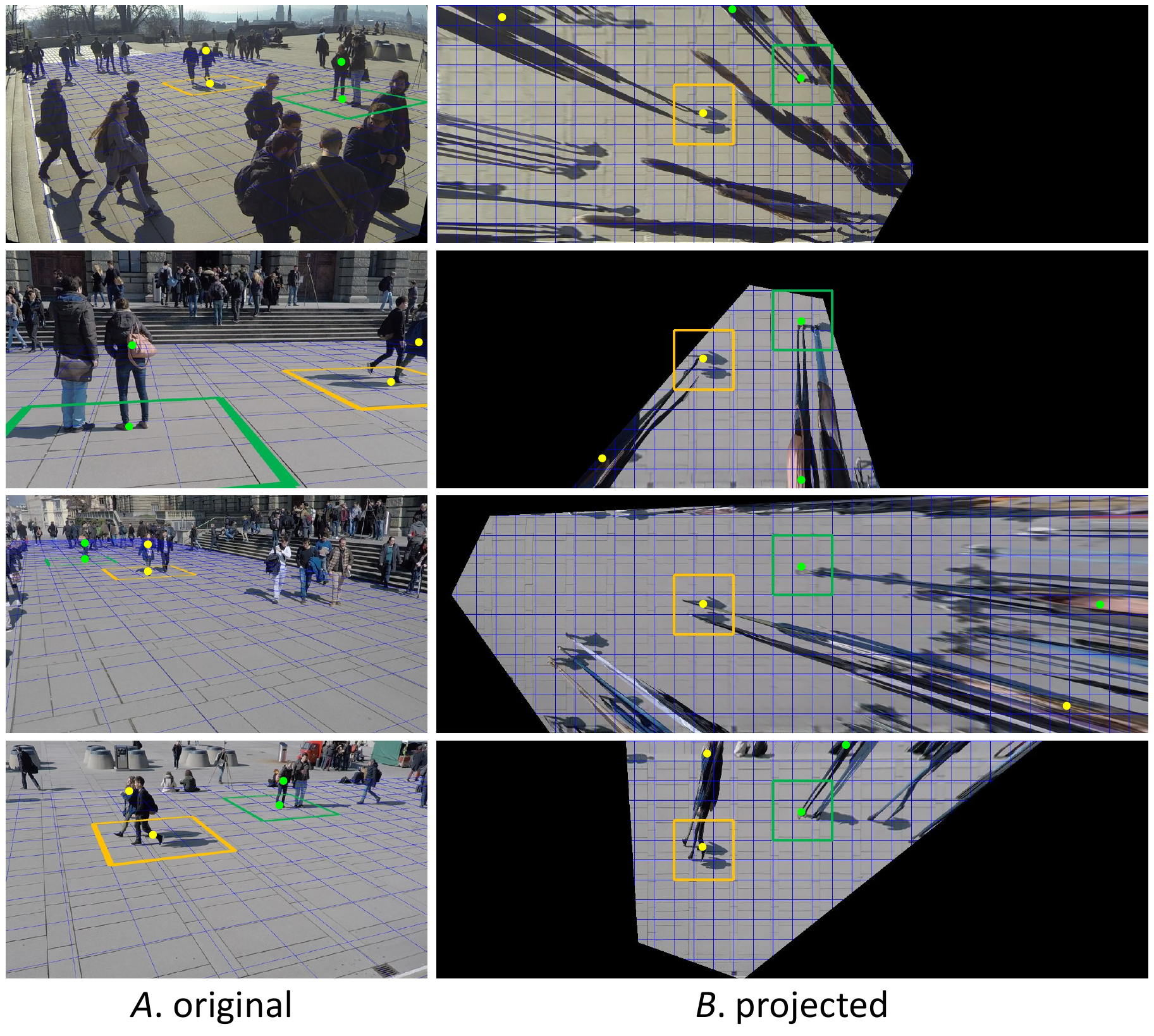}
  \caption{
  Distinct distortion patterns at different locations \textit{vs.} fixed computation in convolution. 
  We highlight two pedestrians with \textcolor{myyellow}{yellow} and \textcolor{mygreen}{green} dots as example. 
  \textit{A}. original images from multiple cameras. \textit{B}. projected images onto the ground plane with bird's eye view, where we observe shadow-like distortion patterns that vary with the locations and views.
  Currently, convolution (\textcolor{myyellow}{yellow} and \textcolor{mygreen}{green}  boxes in \textit{B} and their corresponding receptive fields in \textit{A}) is adopted to aggregate across multiple projected feature maps (we use RGB images here for better illustration). Due to its translation invariance, convolution applies the same calculation at all locations, even when distortion patterns at different positions 
  might be rather different. 
  Therefore, we investigate other design choices for multiview aggregation, which can attend adaptively at different locations, thus better accommodating the shadow-like distortion patterns. 
  }
  \label{fig:intro}
\end{figure}


Incorporating multiple camera views effectively alleviates occlusion~\cite{roig2011conditional,baque2017deep,hou2020multiview}, a challenging problem in the community. For monocular view systems, occlusion prevents the system from discovering targets hidden behind obstacles or other people. 
On the other hand, multiview systems usually leverage multiview images that are synchronized and have overlapping fields of view, and thus complement each other in inferring what is occluded. 
Camera calibrations are also presumably provided in multiview detection, which enable aggregation from multiple views on the ground plane (a common reference plane) of the bird's eye view. 

A central problem in multiview detection is how to aggregate content from multiple camera views. 
A recent state-of-the-art method, MVDet \cite{hou2020multiview}, adopts a fully-convolutional approach. It first projects feature maps onto the ground plane and then applies convolution to jointly consider neighboring locations over multiple camera views. 
The convolution-based fusion approach, being translation-invariant (same calculations regardless of locations and cameras), arguably does not well suit the projected feature maps. In fact, different positions and different cameras in the scenario might introduce diverse distortion patterns during projection, just as different locations and different street lights cause pedestrians to cast various shadows. 
As shown in Fig.~\ref{fig:intro}, pedestrians (and their features) are projected rather dis-similarly across different positions and cameras. 
Therefore, applying the same operation with convolution (due to its innate characteristic of \textit{translation invariance}) potentially contradicts the distinct \textit{position-sensitive} distortion patterns, further limiting the system performance. 

In this paper, we propose MVDeTr, an end-to-end multiview detector that deals with various distortion patterns during multiview aggregation using shadow transformer. Originated from the deformable transformer \cite{zhu2021deformable}, shadow transformer is designed to attend to different pixels across multiple cameras from a certain position and view. 
Departing from deformable transformers that attend within a single image, shadow transformer jointly considers multiple projected feature maps when updating features of each view. 
Specifically, we feed the projected feature maps, a shared ground plane position embedding, and the learned camera embeddings to the deformable transformer, which make the system aware of the object, position, and camera characteristics all at once. 
This view- and position-aware attention mechanism can better address projection distortions (\eg, skews and stretches) rarely observed in natural images. 
Shadow transformer only has an encoder component, and the full system (MVDeTr) adopts a fully connected layer to output the detection result for each ground plane location.

Together with the proposed architecture, a new training scheme is introduced that includes a new view-coherent data augmentation approach and loss terms from related monocular-view detection systems \cite{lin2017focal,law2018cornernet,zhou2019objects}. 
Despite its success in preventing overfitting and allowing higher accuracy, data augmentation is rarely applied in multiview detection systems. This is because of the restriction that the augmented input images must maintain multiview consistency: for joint consideration of multiple camera views to benefit the detection system, these views must be coherent in the content they provide. Since the augmentation policies are mostly random, the augmented input views are likely to violate the consistency restriction, rendering them unusable in multiview systems. 
In the proposed view-coherent data augmentation, we use different random affine transformations
for different views, and extract image features from these augmented images. We also adjust the ground truth accordingly for per-view supervision. Then, we invert the affine transformations before projecting the feature maps onto the ground plane, thus restoring the multiview coherency for ground plane supervision when applying random data augmentations. 


On Wildtrack \cite{chavdarova2018wildtrack}, a real-world dataset, and MultiviewX \cite{hou2020multiview}, a synthetic dataset, MVDeTr achieves state-of-the-art detection accuracy. Specifically, MVDeTr requires a similar number of training iterations and memory footprints compared to the convolution-based alternatives.

\section{Related Work}
\textbf{Monocular-view detection.} The majority of works in object detection focus on the monocular-view setting, where only a single image is available. Many inspiring methods are proposed for this task, including two-stage detectors like the R-CNN family \cite{ren2015faster,girshick2014rich} that regress bounding boxes from proposals, and one-stage detectors that adopt the anchor-based \cite{lin2017focal} or anchor-free \cite{tian2019fcos,zhou2019objects} approach. Some also investigated other detection like point cloud \cite{10.1145/3394171.3413805} and more \cite{10.1145/3394171.3413832,10.1145/3394171.3414427,10.1145/3394171.3413945,10.1145/3394171.3416297,10.1145/3394171.3413816}. 
In addition to the architectures, many novel training schemes are also investigated. To list a few, 
Rezatofighi \etal~\cite{rezatofighi2019generalized} propose a novel localization loss, G-IoU, directly derived from intersection-over-union for object detector training. 
Lin \etal \cite{lin2017focal} propose Focal Loss to deal with the foreground-background imbalance in single-stage detectors. 
Many researchers focus on the pedestrian detection problem, rather than general-case object detection. To deal with occlusions, some investigate part-based methods, as the occluded human are still partially observable~\cite{ouyang2015partial,tian2015deep,noh2018improving,zhang2018occlusion}, while others investigate learned NMS \cite{hosang2017learning}, specific loss design like Repulsion Loss \cite{wang2018repulsion}, and more \cite{10.1145/3394171.3413634,10.1145/3394171.3413989,10.1145/3394171.3413903}.


\textbf{Multiview detection.}
In order to detect pedestrians under heavy occlusion, many researchers choose to explore multiple camera views. These views are synchronized and calibrated, providing complementary descriptions of the scene. 
Note that camera calibrations also produce a correspondence between each location on the ground plane and its bounding boxes in multiple camera views, as we can calculate the 2D bounding boxes once we assume an average 3D human height and width via perspective transformation. In this case, multiview detection systems usually evaluate their performance with pedestrian occupancy maps on the ground plane~\cite{baque2017deep,chavdarova2018wildtrack,hou2020multiview}. 
To aggregate from multiple views, the core problem in multiview detection, researchers adopt different approaches. 
Baque \etal \cite{baque2017deep} use higher-order potentials to model the consistency between spatially neighboring locations, but requires additional potential terms design and calculation outside neural networks. Hou \etal \cite{hou2020multiview} also considers the spatially neighboring locations in multiview aggregation, but accomplish this via convolution and achieve state-of-the-art performance. 
Of all these methods, multiview aggregation either needs specific design outside neural network \cite{fleuret2007multicamera,baque2017deep}, or fail to consider spatial neighboring locations \cite{chavdarova2017deep}, or adopt computations un-adjustable to the locations \cite{hou2020multiview}. 
With that said, none of the current systems applies data augmentations in their training scheme, mainly due to the restriction of multiview coherency.

\textbf{Transformers and their application.}
Recently, researchers investigate transformers \cite{vaswani2017attention,devlin2018bert} and their applications \cite{dosovitskiy2021an,carion2020end,ye2019cross,yang2020learning,sun2019videobert}. 
Originally applied in sequence reasoning tasks, transformers adopt multi-head self-attention to model the relationship between each point. Compared to Non-local networks \cite{wang2018non}, it uses fully connected layers to improve model capability, while providing location sensitivity by feeding position embedding. 
On natural language tasks including machine translation and question answering, transformers achieve great performance, proving their scalability and capability. 
For vision tasks, Dosovitskiy \etal propose ViT as a transformer-based image classifier \cite{dosovitskiy2021an}. In object detection, DETR \cite{carion2020end} use a transformer to model the object relationship in feature maps, and adopt a novel end-to-end object detection training scheme departing from current ones. Given the long training time and large computation overhead of DETR, Zhu \etal propose Deformable DETR, which reduces the computation by only attending to several points around the reference rather than the entire feature map. Similar to Deformable convolution \cite{dai2017deformable}, the attended points in Deformable DETR are also fully learnable.

\section{Method}
In this section, we first revisit the multiview aggregation problem and deformable transformer. Then, we propose the MVDeTr architecture that enables adaptive attention across different locations and cameras. Lastly, we introduce the new training scheme for MVDeTr that allows random data augmentation in multiview detection while maintaining multiview coherency.

\subsection{Revisiting Multiview Aggregation and Deformable Transformer}

\textbf{Multiview aggregation with feature projection.} In the recent state-of-the-art, Hou \etal \cite{hou2020multiview} adopt an anchor free representation, by projecting the multiview feature maps to the ground plane (bird's eye view). Given 2D image pixel coordinate $\left(u,v\right)$, we calculate the 3D world position $\left(x,y,z\right)\big\rvert_{z=0}$ on the ground plane ($z=0$) with 
\begin{equation}
\label{eq:perspective_3x4}
    \gamma\left(\begin{matrix}u \\ v \\ 1\end{matrix}\right) = \bm{P}_\theta\left(\begin{matrix} x \\ y \\ z \\ 1\end{matrix}\right) = \bm{A} \left[\bm{R}|\bm{t}\right] \left(\begin{matrix} x \\ y \\ z \\ 1\end{matrix}\right) = \left[\begin{matrix} \theta_{11} & \theta_{12} & \theta_{13} & \theta_{14} \\ \theta_{21} & \theta_{22} & \theta_{23} & \theta_{24} \\ \theta_{31} & \theta_{32} & \theta_{33} & \theta_{34} \end{matrix}\right] \left(\begin{matrix} x \\ y \\ z \\ 1\end{matrix}\right),
\end{equation}
where $\gamma$ is a scaling factor, and $\bm{P}_\theta$ is the perspective transformation matrix calculated using the intrinsic parameter $\bm{A}$ and rotation-translation matrix (extrinsic parameter) $\left[\bm{R}|\bm{t}\right]$. By setting $z=0$, we can retrieve the correspondence between image pixel $\left(u,v\right)$ and the ground plane position $\left(x,y\right)$, allowing for feature map projection. 

By concatenating all projected feature maps, the system then gains access to a feature description for each location across all cameras. This anchor-free representation outperforms previous anchor-based ones \cite{baque2017deep,chavdarova2017deep} that rely on ROI-pooling \cite{ren2015faster} over the anchor boxes, as anchor boxes might not be accurate in the first place and can lead to inaccurate feature representations. 

Moreover, researchers find that jointly considering neighboring locations with conditional random field (CRF) \cite{baque2017deep} or convolution \cite{hou2020multiview} also helps with the multiview aggregation. But these methods also have their drawbacks: CRF requires specific potential terms design and operations outside neural network forward pass, whereas convolution is translation-invariant and does not suit the various distortion patterns from projection (see Fig.~\ref{fig:intro}). 

\textbf{Deformable transformer.} Proposed in Deformable DETR \cite{zhu2021deformable}, deformable transformer departs from previous self-attention-based transformer architectures \cite{vaswani2017attention} that attend to all locations, and only focuses on a few selected reference points. Specifically, given feature map $\bm{f}$ and a position $\bm{p}$, for each head $m\in\{1,...,M\}$ in the $M$-head pipeline, deformable attention first outputs position offsets $\{\Delta \bm{p}_{mk}\}\big\rvert_{k=1}^K$ for $K$ reference points. Then, it estimates attention weights $\{a_{mk}\}$ for the reference points $\{\Delta \bm{p}_{mk}\}$ in each head ($\sum_{k}{a_{mk}}=1$), and outputs the final results over multi-head,
\begin{equation}
\label{eq:attn}
    \text{DeformAttn}(\bm{f},\bm{p}) = \sum_{m=1}^{M} \bm{W}_m \sum_{k=1}^{K}{a_{mk} \; \bm{W}'_m \; \bm{f}(\bm{p}+\Delta \bm{p}_{mk})},
\end{equation}
where $\bm{W}_m$ and $\bm{W}'_m$ denote the translations for head $m$. 

Such deformable multi-head attention allows for adaptive sampling positions $\{\bm{p}+\Delta \bm{p}_{mk}\}$ for each position $\bm{p}$.
However, it still lacks specific designs so as to aggregate from multiple camera views rather than a single image.

\begin{figure*}[h]
  \centering
  \includegraphics[width=\linewidth]{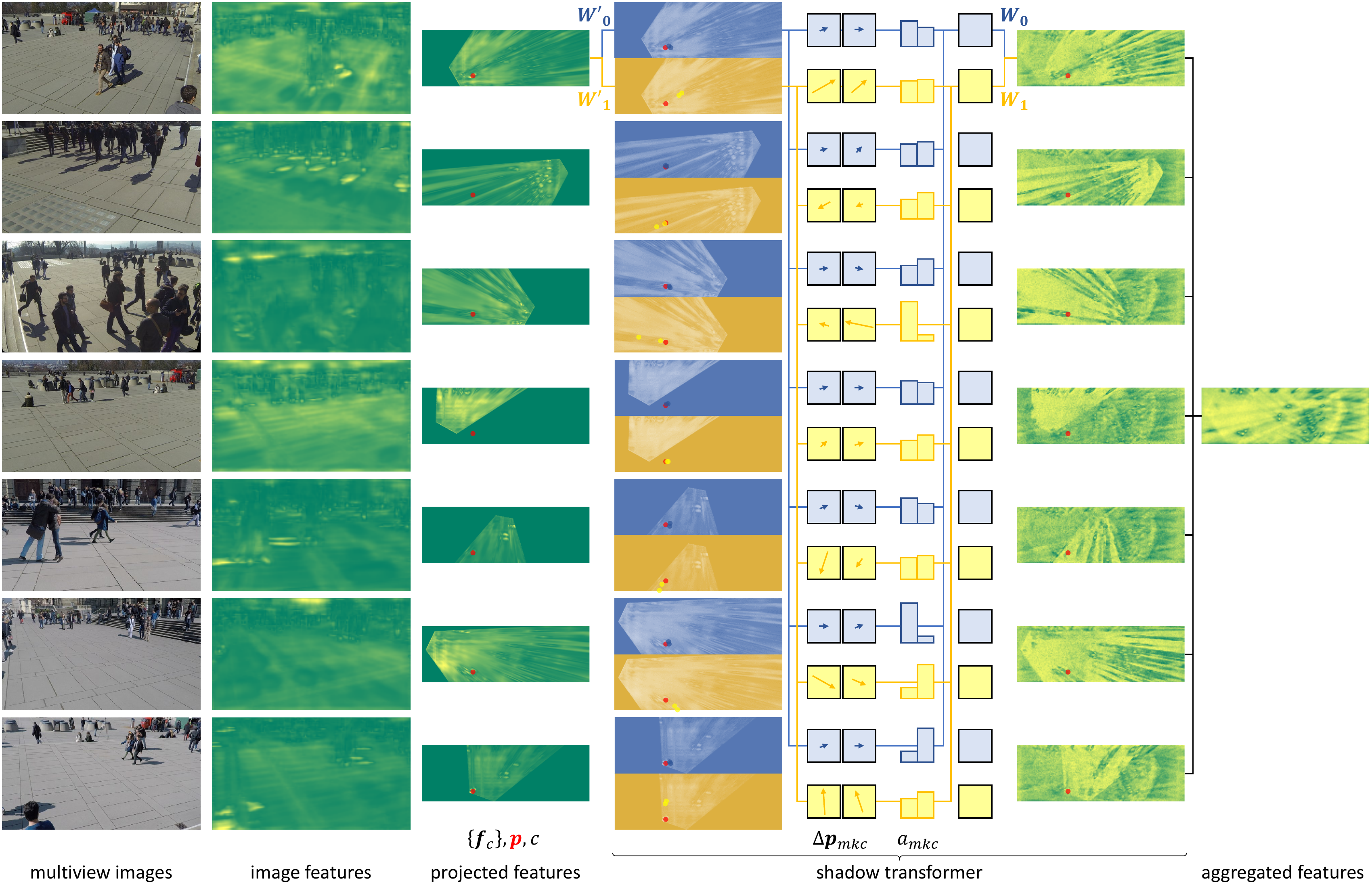}
  \caption{
  Multiview aggregation in MVDeTr. 
  Given input images from $C$ cameras, first, MVDeTr extracts image feature maps. Next, MVDeTr projects the feature maps onto the ground plane. Then, we adopt a shadow transformer to deal with the various shadow-like distortion patterns introduced by the perspective transformation. Last, we aggregate the output feature maps to formulate a final representation for the entire scenario. 
  Specifically, inside this shadow transformer, we use multiview deformable attention (an extension to the deformable attention \cite{zhu2021deformable}) to build an encoder for multiview processing and aggregation.
  To start with, we translate each of the projected feature maps $\{\bm{f}_c\}$ into $M$ heads. After that, at each position $\bm{p}$, we estimate the position offsets of $K$ reference points $\{\bm{p}+\Delta \bm{p}_{mkc}\}$ to attend on, and estimate their corresponding attention weights $\{a_{mkc}\}$ ($\sum_{k,c}{a_{mkc}}=1$). Different from the original deformable transformer design, we consider all different cameras and $M$ heads for each camera (which makes a total number of $M \times K \times C$ reference points for each location in each camera) for the multi-head attention, so as to infuse the multiview information to each of the views.
  }
  \label{fig:agg}
\end{figure*}

\subsection{MVDeTr with Shadow Transformer}
In this paper, we propose MVDeTr, a novel multiview detector that adopts a so-called shadow transformer to aggregate from multiple projection-distorted feature maps (Fig.~\ref{fig:intro}). Rather than a single image as usually does in deformable transformers, shadow transformer jointly considers multiple camera views. 
The overall architecture of MVDeTr is shown in Fig.~\ref{fig:agg}.

Given inputs from multiple camera views, first, MVDeTr extracts image feature maps via a ResNet-18 \cite{he2016deep} network. Next, similar to the previous method \cite{hou2020multiview}, we project the feature maps onto the ground plane to enable further multiview fusion. Then, cues from multiple views are jointly and adaptively processed with the shadow transformer, where a multiview-deformable-attention-based transformer encoder is adopted. We do not use any transformer decoder as we still formulate the multiview detection problem as a key point regression problem \cite{law2018cornernet,zhou2019objects} rather than as a direct set prediction problem \cite{carion2020end} (see Section~\ref{secsec:train} for more details). Lastly, outputs from multiple views are aggregated using a fully connected layer. 
It is noteworthy that the joint consideration of multiple views only relies on the shadow transformer and the last fully connected layer. In this manner, MVDeTr fuses multiview features with \textit{position-sensitive} calculations, in contrast to the \textit{translation-invariant} convolution pipeline that applies the same calculation to all positions. 

Specifically, multiview deformable attention is the core of the proposed shadow transformer. Inspired by multi-scale deformable attention \cite{zhu2021deformable}, we extend deformable attention to multiview detection by jointly considering multiple projected feature maps. 
For $C$ camera views, multiview deformable attention takes as input the projected feature maps $\{\bm{f}_c\}_{c=1}^C$ with shadow-like distortions, a certain position $\bm{p}$ (total number of $\bm{p}$ is the size of the ground plane feature map) across all $C$ feature maps, and learned camera embeddings for each camera. Then, it translates each view into $M$ heads with view-shared fully connected layers, and jointly considers all $M \times K \times C$ reference points when updating for each position at each view, 
\begin{equation}
\label{eq:mvattn}
\begin{aligned}
    & \text{MVDeformAttn}\left(\{\bm{f}_{\hat{c}}\}_{\hat{c}=1}^C,\bm{p},c\right)  \\
=  &\sum_{m=1}^{M} \bm{W}_m \sum_{c'=1}^C{\sum_{k=1}^{K}{a_{mkc'}  \bm{W}'_m \bm{f}(\bm{p}+\Delta \bm{p}_{mkc'})}}.
\end{aligned}
\end{equation}
Specifically, the attention weights satisfy ($\sum_{k,c}{a_{mkc}}=1$).
We use the multiview deformable attention to construct an encoder for multiview processing and aggregation in MVDeTr. 

As shown in Fig.~\ref{fig:agg}, the multi-head architecture allow the system to focus on different aspects. 
For instance, the blue branch focuses on the real shadows cast by the sun, thus attending similar relative positions across multiple views (human shadow cast by the sun are already on the ground plane, and further projection with Eq.~\ref{eq:perspective_3x4} does not bring forward any distortion as shown in Fig.~\ref{fig:intro}).
On the other hand, the yellow branch focuses on the various ``shadow-like'' distortion patterns of human bodies and features, which are caused by perspective transformations at different positions from different cameras
(human bodies are projected onto the ground plane, as if there is a light source at the camera location casting shadows onto the ground plane, thus giving distinct distortion patterns at different locations). Accordingly, the yellow branch attends to different points for different locations. 


\begin{figure*}[h]
  \centering
  \includegraphics[width=\linewidth]{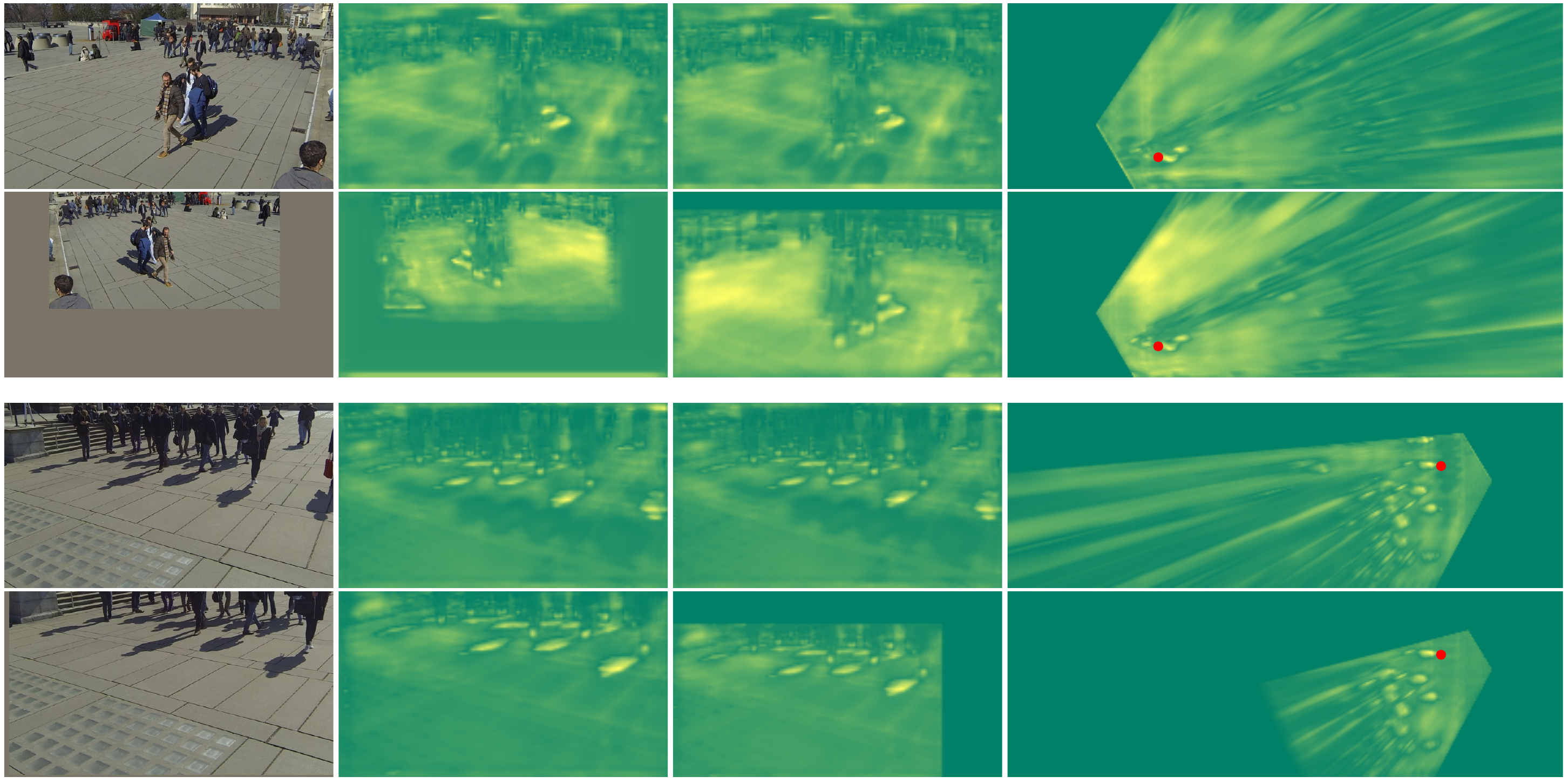}
  \caption{
  View-coherent data augmentation. In each example, we show the original image forward pass on the top and the augmented image forward pass on the bottom. From left to right: input images, image feature maps, un-augmented feature maps, and projected feature maps. Although the projected features are not the same between the original image and the augmented image, the feature distributions around the pedestrian locations (\textcolor{red}{red} points) remain coherent. 
  }
  \label{fig:aug}
\end{figure*}

\subsection{Training Scheme}
\label{secsec:train}
\textbf{Loss functions.}
We formulate multiview detection as a key point detection problem, as the multiview systems evaluate pedestrian occupancy on the ground plane \cite{chavdarova2018wildtrack,hou2020multiview}. 
Following CornerNet \cite{law2018cornernet}, we regress a heatmap of the pedestrian occupancy likelihood score $\hat{s}_{\bm{p}}$ for each position ${\bm{p}}$ on the ground plane. Using Focal Loss~\cite{lin2017focal} and a Gaussian smoothed target $s$ ($s_{\bm{p}}=1$ for the ground truth location and gradually descent to $0$), we write the detection loss as,
\begin{equation}
\label{eq:focal}
    \mathcal{L}_\text{det} = \frac{1}{N}\sum_{\bm{p}}
    \begin{cases}
      (1-\hat{s}_{\bm{p}})^\alpha \log (\hat{s}_{\bm{p}}) & \text{if $s_{\bm{p}}=1$}\\
      (1-s_{\bm{p}})^\beta(\hat{s}_{\bm{p}})^\alpha \log (1-\hat{s}_{\bm{p}}) & \text{otherwise},
    \end{cases}
\end{equation}
where $N$ denotes the total number of pedestrians over all positions. 
Besides, since the heatmap we output is often of lower resolution compared to the ground truth, we also regress an offset $\hat{\bm{o}}_{\bm{p}}$, which aims to compensate for the omitted decimal part $\frac{\bm{p}}{r} - \floor*{\frac{\bm{p}}{r}}$ during the $r\times$ downsampling. The offset loss can be written as,
\begin{equation}
\label{eq:offset}
    \mathcal{L}_\text{off} = \frac{1}{N}\sum_{\bm{p}\rvert_{s_{\bm{p}}=1}}{\left|\bm{o}_{\bm{p}}-\left(\frac{\bm{p}}{r} - \floor*{\frac{\bm{p}}{r}}\right)\right|},
\end{equation}
where $\left|\cdot\right|$ denotes the $\mathcal{L}_1$ norm. 

In addition to the detection loss $\mathcal{L}_\text{det}$ and offset loss $\mathcal{L}_\text{off}$ for the ground plane detection results, we also consider each of the individual images for supervision. 
Specifically, rather than detecting top-left and bottom-right corner \cite{law2018cornernet,duan2019centernet,song2018small} or the object center point \cite{zhou2019objects,duan2019centernet,liu2019high}, we detect the pedestrian foot point and regress its height and width. Foot points, or the center of the bounding box bottom, should be where the bounding box intersects with the ground plane ($z=0$). Therefore, foot points on each image should be the same as or close to the corresponding pedestrian occupancy on the ground plane, and detecting foot points on each image should help to regress the ground plane occupancy. Additionally, estimating bounding box shapes also provides additional supervision, as the ground plane supervision does not include pedestrian height and width. We thus include another bounding box regression loss $\mathcal{L}_\text{box}$ using the same $\mathcal{L}_1$ distance as the offset loss $\mathcal{L}_\text{off}$ over ground truth foot point locations. 

Together, we have the overall loss functions for MVDeTr,
\begin{equation}
\label{eq:loss}
    \mathcal{L} = \mathcal{L}_\text{det}+\mathcal{L}_\text{off}+\frac{1}{C}\sum_{c}{\left( \mathcal{L}_{\text{det},c}+\mathcal{L}_{\text{off},c}+0.1\times \mathcal{L}_{\text{box},c}\right)},
\end{equation}
where $\mathcal{L}_{\cdot,c}$ denotes the image-wise loss for a certain camera $c$.

\textbf{View-Coherent Data Augmentation.}
Data augmentation helps to prevent the system from overfitting. However, this is non-trivial as multiview systems must maintain view consistency after augmentation. In this work, we design a view-coherent data augmentation scheme to meet such requirements. 

First, for each input view, we apply random data augmentations including horizontal flipping, cropping, and scaling. Specifically, we record the affine matrix for the random transformations applied during data augmentation. 

Next, we extract features from such augmented images. The per-view supervisions are also adjusted accordingly to the augmentation applied to the image for pedestrian detection.

Then, we apply the inverse affine transformation to create un-augmented feature maps (see Fig.~\ref{fig:aug}). Such un-augmented feature maps do not equal the feature maps of the original image forward pass, but maintain the pedestrian locations and have similar feature distributions around these locations. 

Finally, we project the un-augmented image features to the ground plane, using the same perspective transformations with the camera pose and calibration. In this manner, we apply random data augmentation to the multiview images without breaking the view-consistency requirements.

\begin{table*}[]
\centering
\begin{tabular}{l|cccc|cccc}
\toprule
                   & \multicolumn{4}{c|}{Wildtrack} & \multicolumn{4}{c}{MultiviewX} \\ \hline
Method             & MODA  & MODP  & Precision & Recall & MODA  & MODP  & Precision  & Recall \\ \hline
RCNN \& clustering \cite{xu2016multi} & 11.3  & 18.4  & 68    & 43     & 18.7  & 46.4  & 63.5   & 43.9   \\ 
POM-CNN \cite{fleuret2007multicamera}           & 23.2  & 30.5  & 75    & 55     & -     & -     & -      & -      \\ 
DeepMCD \cite{chavdarova2017deep}           & 67.8  & 64.2  & 85    & 82     & 70.0  & 73.0  & 85.7   & 83.3   \\ 
Deep-Occlusion \cite{baque2017deep}    & 74.1  & 53.8  & 95    & 80     & 75.2  & 54.7  & 97.8   & 80.2   \\ 
MVDet \cite{hou2020multiview}             & 88.2  & 75.7  & 94.7  & 93.6   & 83.9  & 79.6  & 96.8   & 86.7   \\ \hline
MVDet (w/ augmentation)             & 89.0  & 75.5  & 93.5  & 95.5   & 85.6  & 78.0  & 96.4   & 89.2   \\ 
MVDeTr (deform attention)             & 90.2  & 81.6  & 96.9  & 93.2   & 93.0  & 91.1  & 99.4   & 93.6   \\ 
MVDeTr (w/o augmentation)             & 89.5  & 81.2  & 94.0  & \textbf{95.6}   & 93.1  & 90.9  & 98.5   & \textbf{94.4}   \\ 
MVDeTr (w/o per-view loss)             & 89.9  & 81.8  & 97.7  & 92.1   & 92.6  & 90.6  & 99.3   & 93.3   \\ \hline
\textbf{MVDeTr (ours)}             & \textbf{91.5}  & \textbf{82.1}  & 97.4  & 94.0   & \textbf{93.7}  & \textbf{91.3}  & 99.5   & 94.2   \\ \bottomrule
\end{tabular}
\caption{Performance (\%) comparison on Wildtrack and MultiviewX datasets. ``MVDeTr (ours)'' is the  proposed method. ``MVDeTr (convolution)'', ``MVDeTr (deform attention)'', ``MVDeTr (w/o augmentation)'', and ``MVDeTr (w/o per-view loss)'' are variants of the proposed method. See Section \ref{secsec:ablations} for more details. }
\label{tab:sota}
\end{table*}

\begin{figure*}[h]
  \centering
  \includegraphics[width=0.9\linewidth]{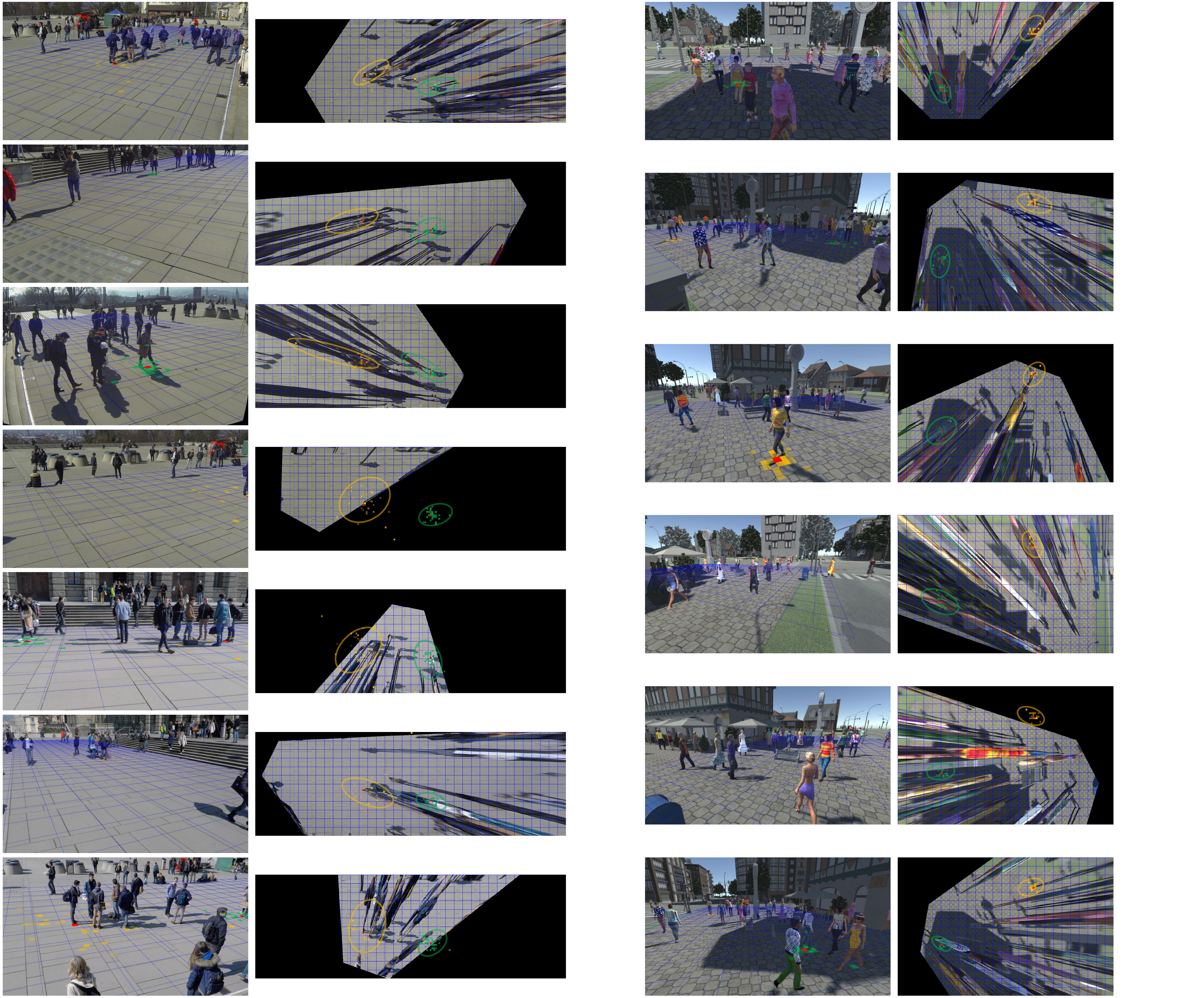}
  \caption{Visualization of the attended points and their corresponding weights in MVDeTr. We show the results on Wildtrack and MultiviewX datasets. For each dataset, we show the original image on the left, and the projected image onto the ground plane on the right. The learned attentions and attended positions on certain positions (\textcolor{red}{red} points) are shown with \textcolor{myyellow}{yellow} or \textcolor{mygreen}{green} points of different shades. We circle out the most prominent points (with high attention weights) in each projected view. The shadow transformer in MVDeTr learns to attend to different points at different positions and cameras, demonstrating its effectiveness in dealing with the various shadow-like distortion patterns. 
  }
  \label{fig:visual}
\end{figure*}

\section{Experiments}
\subsection{Experiment Settings}
\textbf{Datasets.} We verify the performance of the proposed detector and training scheme on the following datasets.

\textit{Wildtrack} \cite{chavdarova2018wildtrack} is a real-world dataset that focuses on a $12 \times 36$ square meters region with 7 synchronized cameras. The ground plane is discretized into and annotated at a resolution of $480 \times 1440$, and the images are of $1080 \times 1920$ pixels. It consists of 400 frames, with the first 360 frames dedicated for training and the last 40 frames used for testing. 

\textit{MultiviewX} \cite{hou2020multiview} is a synthetic dataset created using the Unity \cite{unity} engine. It records a $16 \times 25$ square meters city square with 6 cameras, while having almost double the pedestrian crowdedness compared to the Wildtrack dataset to compensate for the limited appearance variance in the synthetic engine. Its ground plane is discretized into a $640 \times 1000$ grid, and the images are also taken with a $1080 \times 1920$ resolution. Similarly, it also has 400 frames with the last 40 frames used for evaluation. 

\textbf{Evaluation metrics.}
Different from monocular-view detection systems that evaluate the estimated bounding boxes, multiview detection systems evaluate the estimated ground plane occupancy map. Specifically, rather than intersection-over-union (IOU) between the estimated and ground truth bounding boxes, multiview detection systems consider the distance between the estimated pedestrian location and its ground truth, and use a 0.5-meter threshold for deciding true positives. 
Following previous works \cite{chavdarova2018wildtrack,hou2020multiview}, we use multi-object detection accuracy (MODA) as the primary indicator, since it considers both false positives and false negatives.

\textbf{Implementation details.}
We first downsample all input images to $720 \times 1280$ for all $C=7$ or $C=6$ cameras since all images need to be considered at the same time. 
Inside the ResNet-18 feature extractor, we reduce the feature map downsample rate to 8 by replacing the last 2-strided convolutions with dilated convolutions, so as to increase its size while maintaining the receptive field. In this manner, we can preserve a relatively high-definition feature map, which benefits the detection system. To further reduce the memory footprint, we add a 128 channel bottleneck to the ResNet-18 feature extractor, before the per-view output heads and the feature projection module. 
For feature map perspective transformation and the ground plane heatmap supervision, we downsample the world grid by a factor of $r=4$. 
For the shadow transformer, we build the transformer encoder for multiview processing and fusing with 3 encoder layers, each has $M=8$ heads and attends to $K=4$ points in each head. The shadow transformer also takes an additional $2\times$ downsampling on the projected ground plane feature maps due to memory concerns. 

During training, we randomly apply horizontal flipping, cropping, and scaling with the proposed view-coherent augmentation. We train the detector using an Adam optimizer \cite{kingma2015adam} with one-cycle learning rate scheduler \cite{smith2019super} at a maximum learning rate of $1\times10^{-3}$. We initialize the ResNet-18 using the ImageNet \cite{imagenet_cvpr09} pre-trained model and set its learning rate to $0.1\times$ of all other layers. 
All experiments are conducted using one RTX-2080TI GPU.

\subsection{Evaluation of MVDeTr}
In Table \ref{tab:sota}, we compare the multiview detection performance of the proposed method against previous methods. Specifically, we find that MVDeTr outperforms the previous state-of-the-art by $+3.3\%$ MODA on the Wildtrack dataset, and $+9.8\%$ MODA on the MultiviewX dataset. We point out that such improvements bring the MODA performance of MVDeTr to relatively high accuracies of  $91.5\%$ and $93.7\%$ MODA on the two datasets.

In Fig.~\ref{fig:visual}, we visualize the learned multiview deformable attention. All attended points ($M \times K \times C$ reference points in Eq.~\ref{eq:mvattn}) of the same color are from a certain point of a certain camera. When considering the points that contribute mostly (in each circle), we find that the attention displays distinct patterns at different positions and different camera, while considering both their neighbors and the projected human bodies. Such adaptive attention patterns well align with our goal to deal with the various shadow-like distortion patterns, verifying the effectiveness of the proposed method. 

\begin{figure*}[h]
  \centering
  \includegraphics[width=\linewidth]{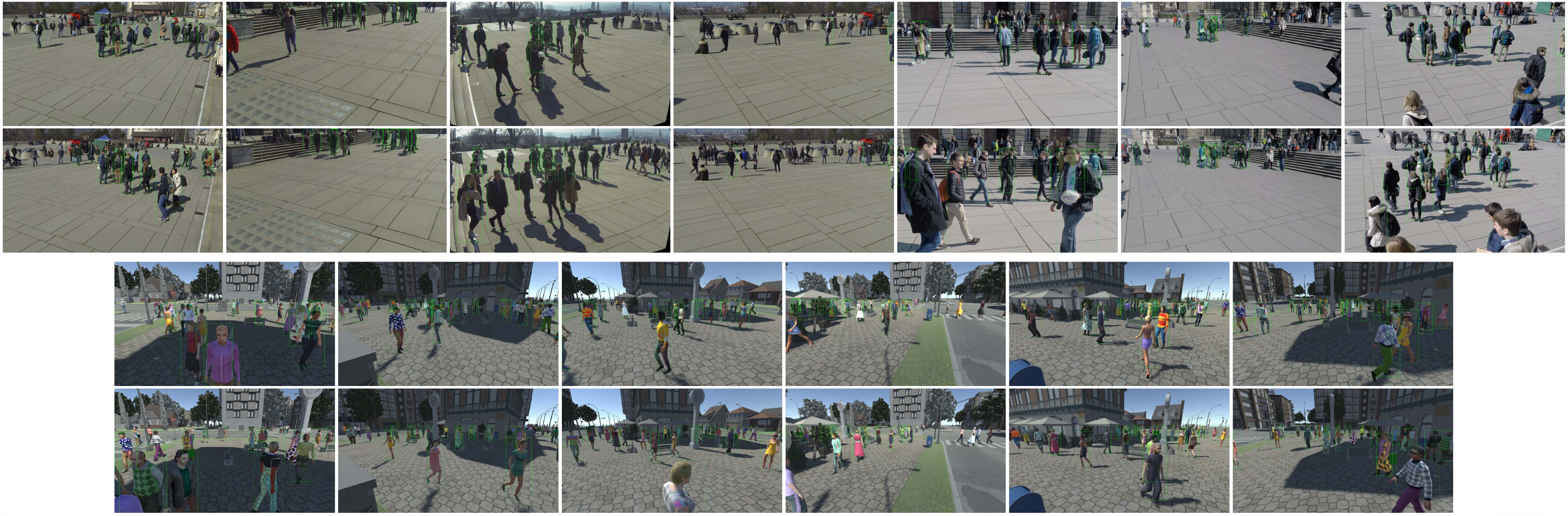}
  \caption{Detection result visualization of the proposed method. 
  }
  \label{fig:results}
\end{figure*}

We further visualize the detection results from the proposed method in Fig.~\ref{fig:results}, where most pedestrians are well detected even though the scene is heavily crowded and most pedestrians are occluded in certain views.

\begin{figure}[h]
  \centering
  \includegraphics[width=0.8\linewidth]{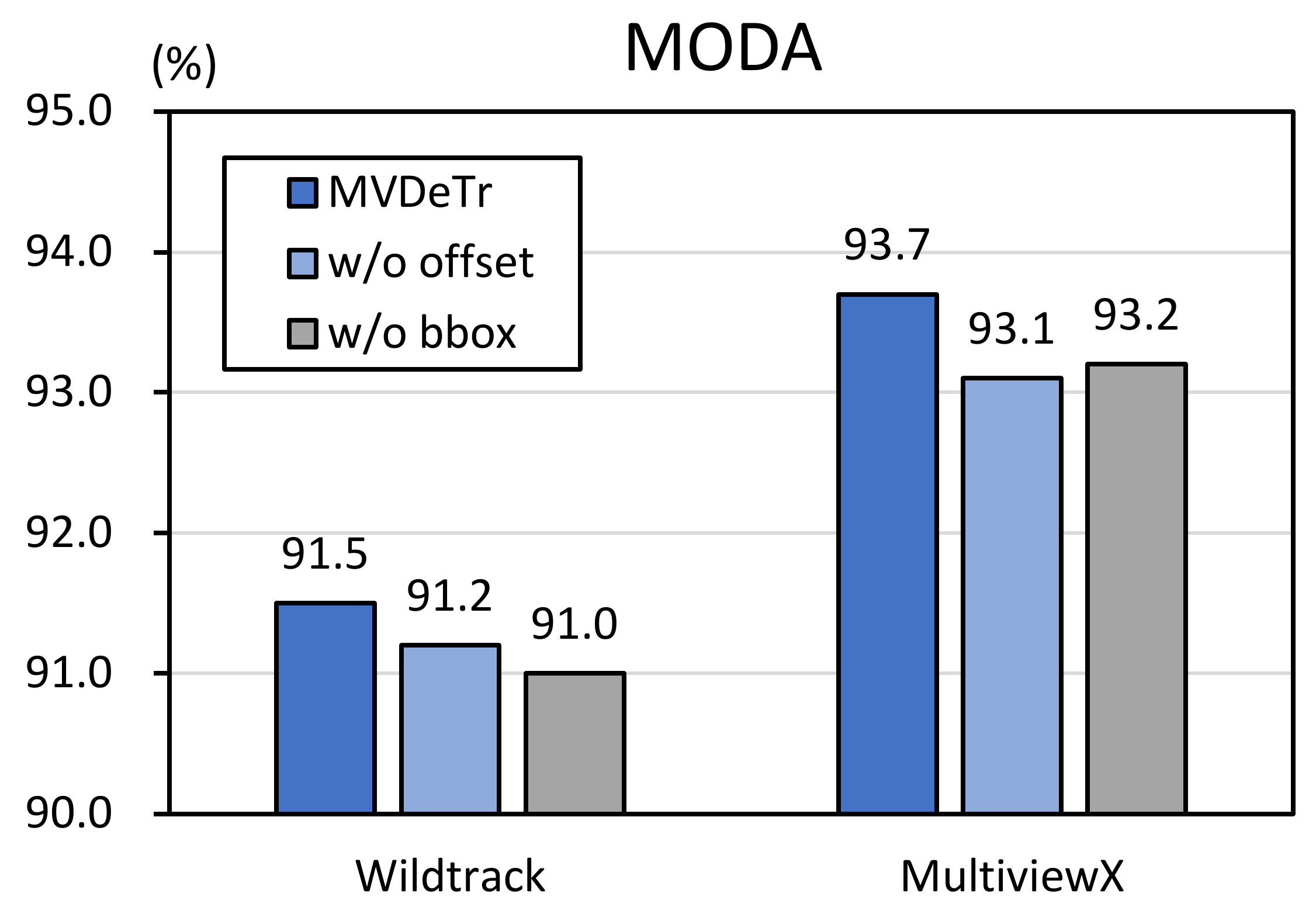}
  \caption{Influence of different loss terms.
  }
  \label{fig:loss}
\end{figure}

\subsection{Ablations}
\label{secsec:ablations}

\textbf{Improvements over convolution.}
In order to quantitatively compare the performance difference between convolution-based multiview aggregation and the proposed shadow transformer alternative, we evaluate the convolution variant ``MVDeTr (convolution)'' in Table~\ref{tab:sota}. ``MVDeTr (convolution)'' is very similar to the previous method MVDet \cite{hou2020multiview}. Their main differences are that the former has a few modifications (e.g., a 128-dim bottleneck, different output heads) to fit it into one RTX 2080TI GPU, and adopts the same training scheme as the full MVDeTr. 
Replacing multiview deformable attention with convolution leads to $-2.0\%$ and $-1.0\%$ MODA losses on Wildtrack and MultiviewX, respectively. The raw MODA differences are relatively small, but we point out that they are over some very high baselines (the convolution variant) and are statistically very significant (\ie, \textit{p}-value~$<$~0.001 over 5 runs). 
Although this variant achieves higher precisions, its lower recalls prevent it from outperforming the proposed method. We believe the main reason for this is that the convolution approach applies the same calculation over entire projected feature maps, which can provide slightly more confident results (higher precision) but often at the price of failing to detect at some uncertain positions (much lower recall). 

\textbf{Improvements over deformable attention.}
We show the effectiveness of our multiview deformable attention over deformable attention \cite{zhu2021deformable} in Table \ref{tab:sota}. Specifically, we replace all multiview deformable attention with deformable attention in our shadow transformer, and create the ``MVDeTr (deform attention)'' variant. In this manner, the multiview detector only fuses from the last fully connected layer before the output heads. 
When compared against the proposed method, ``MVDeTr (deform attention)'' falls short by $-1.7\%$  MODA on Wildtrack and $-0.7\%$ MODA on MultiviewX. Again, such performance differences are small in raw values but statistically significant, with \textit{p}-value~$<$~0.05 over 5 runs. 
This also verifies that our multiview deformable attention design is more effective for multiview detection, as it jointly considers multiple views to process and fuse.

\textbf{Effectiveness of the view-coherent data augmentation.}
In Table~\ref{tab:sota}, removing data augmentation from MVDeTr causes a $-2.0\%$ MODA loss on Wildtrack dataset and a $-0.6\%$ MODA loss on MultiviewX dataset. 
Adding the proposed view-coherent data augmentation also helps the MVDet system, and improves its MODA by $+0.8\%$ and $+1.7\%$ on two datasets. 
Overall, data augmentation gives better precision (fewer false positives) and worse recall (more false negatives). The system trained with data augmentation is less likely to consider an unsure position occupied by a pedestrian as it predicts fewer positives. Data augmentation might crop out a certain part of the image, which can create more unsure candidates. As the ground truth positives (on the ground plane) remain the same, more unsure candidates during training can reduce the tendency of considering unsure positions as occupied during testing, further leading to fewer predicted positives. On the whole, as the trade-off between fewer false positives and more false negatives still leads to higher MODA, we consider the view-coherent augmentation beneficial to the overall system.


\textbf{Effectiveness of the per-view loss.}
As shown in Table~\ref{tab:sota}, the removal of the per-view loss results in MODA drops of $-1.6\%$ and $-1.1\%$ on Wildtrack and MultiviewX, respectively. This aligns with previous findings by Hou \etal \cite{hou2020multiview}. In fact, we believe that despite being formulated as a ground plane key-point detection problem, multiview detection can still benefit from the supervision of un-projected images. Especially, with the help of the view-coherent data augmentation, our correspondingly adjusted per-view supervision can also further prevent the system from overfitting.

\textbf{Influence of the offset and bounding box loss terms.}
In Fig.~\ref{fig:loss}, we further investigate the influence of the offset loss and the bounding box loss. Specifically, we find the removal of either of these leads to a performance drop on both datasets. Offset loss is slightly more important for the MultiviewX dataset (the removal of it causes a $-0.6\%$ MODA loss compared to $-0.3\%$ on Wildtrack), as the synthetic human can sometimes move too close together and the absence of the offset estimation can lead to potentially more false negatives from non-maximum suppression. Bounding box regression, on the other hand, is similarly important for both datasets (the removal of it results in $-0.5\%$ MODA loss for both datasets), indicating that regressing the bounding box shapes is beneficial to the system training.

\section{Conclusion}
In this paper, we investigate how to jointly consider multiple views to combat the occlusions in detection. Specifically, we find that convolution might not be effective for multiview feature aggregation under various \textit{position-sensitive} distortions due to its \textit{translation-invariant} nature. In this case, we propose MVDeTr, a multiview detector that adopts shadow transformer for multiview feature fusion. Shadow transformer is designed specifically for the shadow-like distortions, and can attend accordingly at different positions over multiple camera views. Together with MVDeTr, we adopt a new training scheme that includes view-coherent data augmentation, which preserves multiview consistency after random augmentation. 
Using the novel training scheme and MVDeTr, we report new state-of-the-art on multiview detection benchmarks. 

\section*{Acknowledgement}
This work was supported by the ARC Discovery Early Career Researcher Award (DE200101283) and the ARC Discovery Project (DP210102801).

\bibliographystyle{ACM-Reference-Format}
\bibliography{reference}

\end{document}